%% file: main.tex
\documentclass[10pt,sigconf,screen]{acmart}
\usepackage{xspace}
\usepackage{siunitx}
\usepackage{subfig}

\usepackage{amsmath}
\usepackage{amssymb}
\usepackage{rotating}
\usepackage{booktabs}
\usepackage{listings}
\usepackage{bbm}

\usepackage[ruled]{algorithm2e} % For algorithms

\AtBeginDocument{%
  \providecommand\BibTeX{{%
    \normalfont B\kern-0.5em{\scshape i\kern-0.25em b}\kern-0.8em\TeX}}}

\begin{document}
\newcommand{\sys}{\texttt{LiveTune}\xspace}

%%
%% The "title" command has an optional parameter,
%% allowing the author to define a "short title" to be used in page headers.
\title{\sys: Dynamic Parameter Tuning for Feedback-Driven Optimization}

%%
%% The "author" command and its associated commands are used to define
%% the authors and their affiliations.
%% Of note is the shared affiliation of the first two authors, and the
%% "authornote" and "authornotemark" commands
%% used to denote shared contribution to the research.
% \author{Ben Trovato}
% \authornote{Both authors contributed equally to this research.}
% \email{trovato@corporation.com}
% \orcid{1234-5678-9012}
% \author{G.K.M. Tobin}
% \authornotemark[1]
% \email{webmaster@marysville-ohio.com}
% \affiliation{%
%   \institution{Institute for Clarity in Documentation}
%   \streetaddress{P.O. Box 1212}
%   \city{Dublin}
%   \state{Ohio}
%   \country{USA}
%   \postcode{43017-6221}
% }

\newcommand{\nojan}[1]{\textcolor{green}{{\sf (NS:} {\sl{#1})}}}
\newcommand{\aiden}[1]{\textcolor{red}{{\sf (AT:} {\sl{#1})}}}
\newcommand{\soheil}[1]{\textcolor{blue}{{\sf (SZ:} {\sl{#1})}}}

\author{Soheil Zibakhsh Shabgahi, Nojan Sheybani, Aiden Tabrizi, Farinaz Koushanfar}
\affiliation{\textit{University of California, San Diego
}\country{USA} \\
\{szibakhshshabgahi, nsheybani, atabrizi, fkoushanfar\}@ucsd.edu}
% \author{Anonymous Authors}

%%
%% By default, the full list of authors will be used in the page
%% headers. Often, this list is too long, and will overlap
%% other information printed in the page headers. This command allows
%% the author to define a more concise list
%% of authors' names for this purpose.
% \renewcommand{\shortauthors}{Trovato and Tobin, et al.}

%%
%% The abstract is a short summary of the work to be presented in the
%% article.
\begin{abstract}
% Traditional machine learning training is a static process that lacks real-time adaptability of hyperparameters. Popular tuning solutions during runtime involve checkpoints and schedulers. Adjusting hyper-parameters usually require the program to be restarted, wasting utilization and time, while placing unnecessary strain on memory and processors. We present \emph{LiveTune}, a new framework allowing real-time parameter tuning during training through \emph{LiveVariables}. Live Variables allow for a continuous training session by storing parameters on designated ports on the system, allowing them to be dynamically adjusted. Extensive evaluations of our framework show saving up to 60 seconds and 5.4 Kilojoules of energy per hyperparameter change.
% Traditional machine learning methods are fixed and lack the ability to change in real-time. Current solutions, like checkpoints, are inefficient, consuming excessive time and power for restarting the process. We intruduce LiveTune, real-time tuning during training. This is achieved using 'Live Variables', which store parameters on specific system ports. These can be adjusted without restarting the program, ensuring continuous training. Our extensive tests show that LiveTune dramatically lowers CPU and GPU use, outperforming classical training paradigms.
Feedback-driven optimization, such as traditional machine learning training, is a static process that lacks real-time adaptability of hyperparameters. Tuning solutions for optimization require trial and error paired with checkpointing and schedulers, in many cases feedback from the algorithm is overlooked. Adjusting hyperparameters during optimization usually requires the program to be restarted, wasting utilization and time, while placing unnecessary strain on memory and processors. We present \emph{LiveTune}, a novel framework allowing real-time parameter adjustment of optimization loops through \emph{LiveVariables}. Live Variables allow for continuous feedback-driven optimization by storing parameters on designated ports on the system, allowing them to be dynamically adjusted. Extensive evaluations of our framework on standard machine learning training pipelines show saving up to 60 seconds and 5.4 Kilojoules of energy per hyperparameter change. We also show the feasibility and value of LiveTune in a reinforcement learning application where the users change the dynamics of the reward structure while the agent is learning showing $5\times$ improvement over the baseline. Finally, we outline a fully automated workflow to provide end-to-end, unsupervised feedback-driven optimization.
% \nojan{Where are we outlining this?}.

% performing several other feedback-driven optimization techniques that benefit from LiveTune, including reinforcement learning and computer aided design (CAD) optimization. Finally, we outline a fully automated workflow to provide end-to-end, unsupervised feedback-driven optimization.

\end{abstract}

%%
%% The code below is generated by the tool at http://dl.acm.org/ccs.cfm.
%% Please copy and paste the code instead of the example below.
%%


%%
%% Keywords. The author(s) should pick words that accurately describe
%% the work being presented. Separate the keywords with commas.

% \received{20 February 2007}
% \received[revised]{12 March 2009}
% \received[accepted]{5 June 2009}

%%
%% This command processes the author and affiliation and title
%% information and builds the first part of the formatted document.
\maketitle

\input{0_intro}
\input{1_prelims}

\input{2_related}
\input{3_method}
\input{4_eval}

\input{5_conclusion}

\bibliographystyle{ACM-Reference-Format}
\bibliography{refs}

\end{document}

%% file: 0_intro.tex
\section{Introduction}

% \nojan{most of this needs to be redone to make it more research-y. soheil you should try and i'll help too} The field of machine learning is in a state of constant evolution, with the rapid development of new algorithms, models, and techniques. As a result, developers face the persistent challenge of selecting and tuning hyperparameters during the training of machine learning models, a process that can profoundly impact model performance. Traditional methods for hyperparameter tuning, including manual trial-and-error and automated techniques like grid search or random search, have served as the primary means of tackling this challenge.

The decade has seen a surge in feedback-driven optimization, most notably in machine learning paradigms. The increasing need for computational power for fine-tuned optimization in complex systems is pressing, calling for optimized utilization of computing resources.
As one of the most prominent methods of feedback-driven optimization, deep learning's rise is attributed to two primary factors: the availability of vast data sets, and advancements in hardware which enhance computational capabilities. This also applies to several other learning paradigms, such as reinforcement learning.

Recent AI trends involve complex networks with billions of parameters, requiring extended training periods, some of which last months \cite{yu2020hyper}\cite{tan2019efficientnet}. Training the contemporary sophisticated networks demands a careful selection of hyperparameters. Hyperparameters, distinct from model parameters updated during training, are often static or follow predetermined trajectories. They include structural elements like layer counts and per-layer parameters, as well as algorithmic settings such as learning rate, momentum, regularizations, and batch size \cite{yu2020hyper}. The choice of hyperparameters significantly impacts model performance and training speed. Hence, selecting the right hyperparameters requires expertise and deliberate planning before training begins \cite{rodriguez2018understanding}. In common practice, researchers do extensive tests on a smaller subset of the data to determine the neighborhood of the best-performing hyperparameters. Often, this selection involves a thorough exploration of the hyperparameter space. 

Existing methods for hyperparameter tuning, such as grid search \cite{shekar2019grid} and Bayesian optimization \cite{joy2016hyperparameter}, are impractical for ML development and rapid testing. In the experimentation phase of machine learning workflows, manual tuning and testing of different combinations of hyperparameters based on feedback, like the loss curve, is key. Manual tuning is a resource-intensive process that involves setting checkpoints, and then loading and remaking those checkpoints after every variable change. This is a repetitive task and often takes multiple attempts before the neighborhood of the right hyperparameter is found, wasting not only time but also exhausting hardware resources on every restart. After the range of correct hyperparameter values is found, an exhaustive search is done to find the best-performing set of hyperparameters in the specific problem.

The repetitive overhead of prominent learning paradigms is further characterized by the intense and iterative optimization workflow of reinforcement learning (RL). RL requires users to directly incorporate immediate feedback into the training workflow to achieve an optimal policy. In most scenarios, this requires complex algorithms that automate the integration of gathered feedback, or the user is required to stop training and start again with adjusted hyperparameters. In current workflows, users are not able to dynamically adjust RL parameters based on the performance of an agent's current policy, which drastically raises the amount of time and resources needed to find an optimal policy.

In response to the current inefficient workflows for feedback-driven optimization, we propose \sys, a novel framework designed for real-time dynamic adjustment of hyperparameters during runtime. This approach introduces unprecedented flexibility in optimization workflows, offering a new paradigm in continuous tuning while incorporating feedback. The framework contributes to reductions in execution time and power consumption across a wide range of optimization paradigms.

The core of \sys is implemented through our novel ``LiveVariables''. Each LiveVariable instance is initialized with a tag, an initial value, and a designated port on the host machine. Its internal value can be modified through this port. LiveVariables can replace normal variables in any code, such as those representing hyperparameters, allowing developers to update them and see the change immediately, without having to restart the process.
% \sys operates by assigning parameters to dedicated listener threads. These threads continuously monitor specific TCP ports on the host machine. To streamline user experience, \sys initializes a bookkeeping server at the start of the program. This server, referred to as the ``dictionary'' thread, tracks all variable tags and their corresponding ports. The only requirement for developers is to keep track of the program's dictionary port. For seamless integration and compatibility, LiveVariables function as standard variables. When a concrete value is required, invoking the instance returns its current internal value.
Another component of \sys is the ``LiveTriggers''. These are boolean flags that can be embedded in the program code or within loops. Leveraging \sys's dynamic variable modification capability, LiveTriggers can activate or halt procedures based on developer commands, without needing to terminate the program. Upon activation, a LiveTrigger returns `True' once before reverting to its default `False' state.

Utilizing \sys has demonstrated a notable acceleration in optimization processes. This paper will detail the underlying mechanisms of \sys, its applications, and provide empirical evidence supporting its efficacy as a transformative tool in feedback-driven optimization.

In summary, our contributions are as follows:
\begin{itemize}
    \item Introduction of \sys, an innovative end-to-end framework for feedback-driven optimization. It enables real-time, manual hyperparameter tuning in a manner that is energy-efficient and minimally disruptive to the optimization process. 
    \item Versatile design of \sys so that it is applicable to any optimization pipeline regardless of objective or scale while inducing minimal overhead.
    \item Development of ``LiveVariables'' and ``LiveTriggers'', allowing for dynamic variable updates and control over subprocesses without program termination.
    % \item Devising an open source API for \sys \footnote{\url{https://anonymous.4open.science/r/LiveTune-E853/README.md}} to facilitate automation and adaptation of the method in optimizing real-world systems.
    \item Devising an open source API for \sys \footnote{\url{https://github.com/soheilzi/LiveTune}} to facilitate automation and adaptation of the method in optimizing real-world systems.
    \item Extensive evaluations of \sys's open-source API which demonstrates significant time and energy savings compared to conventional optimization in emerging learning paradigms.
\end{itemize}

%% file: 1_prelims.tex
\section{Preliminaries}

\subsection{Hyperparameter Tuning}

Tuning hyperparameters has proven to be the most important task in ML training, as every model architecture and dataset has a unique set of hyperparameter configurations that enable optimal performance \cite{bardenet2013collaborative}. Rather than model parameters, which are learned during training, hyperparameters dictate the learning process and must be iteratively tuned to progressively enable better learning. Some fundamental hyperparameters include learning rate, regularization coefficients, number of epochs, and momentum.

Hyperparameter tuning involves systematically searching for an optimal combination of hyperparameters that results in the best performance of the model. A naive approach to this is a user performing a brute-force search across the entire hyperparameter space. Autonomous methods, such as grid search \cite{shekar2019grid}, allow users to specify a predefined range within the hyperparameter space that they believe will enable high performance. Random search algorithms \cite{mantovani2015effectiveness} randomly sample the hyperparameter space and return the hyperparameters that showed the most promise. Search algorithms are often very resource-intensive and do not guarantee suitable results. More advanced methods, such as Bayesian optimization \cite{joy2016hyperparameter}, use probabilistic models to predict how different hyperparameter configurations will perform. These methods are more efficient than search algorithms, however, they still may require fine-tuning by a user to achieve optimal results.

Even after sophisticated and resource-intensive algorithms are run to find suitable hyperparameter configurations, a human expert is often required to perform their own manual search to pinpoint the ideal hyperparameters for training. This takes a heavy toll on computing time and overhead, as training is in a constant cycle of starting to observe the effects of the hyperparameters and stopping if the hyperparameters are not suitable. \sys addresses this issue by allowing users to update hyperparameters without reloading the program instance. This enables a much more time and resource-efficient approach to training large ML models.

\subsection{Reinforcement Learning}
Reinforcement learning (RL) is a computational paradigm where a learner, called an agent, interacts with a dynamic environment to achieve certain goals by learning to make optimal decisions. The agent receives feedback in the form of rewards and modifies its strategy, termed as policy, to improve its performance over time.

In the RL context, the interaction between the agent and the environment is modeled as a Markov Decision Process (MDP). An MDP provides a mathematical framework for modeling decision-making in situations where outcomes are partly random and partly under the control of a decision maker. It is defined by the tuple $(\mathcal{S}, \mathcal{A}, \mathcal{P}, r)$, where:
\begin{itemize}
    \item $\mathcal{S}$ represents the set of all possible states of the environment.
    \item $\mathcal{A}$ denotes the set of all actions the agent can take.
    \item $\mathcal{P}$ is the transition probability that determines the likelihood of moving from one state to another state given an action. Specifically, $\mathcal{P}(s'|s,a)$ is the probability of transitioning to state $s'$ from state $s$ after taking the action $a$.
    \item $r(s,a)$ is the reward function, which gives the immediate reward received after transitioning from state $s$ to state $s'$ due to action $a$.
\end{itemize}

The objective in RL is to learn a policy $\pi$, a mapping from states to actions, that maximizes the expected cumulative reward. The cumulative reward is often discounted by a factor $\gamma \in [0,1)$, known as the discount factor, which represents the difference in importance between future rewards and immediate rewards. The expected return from a policy $\pi$, starting from state $s$, is given by:
\begin{equation*}
    J_{\pi}(s) = \mathbb{E}\left[\sum_{t=0}^{\infty} \gamma^t r(s_t, a_t) \mid s_0 = s\right]
\end{equation*}
where $s_t$ and $a_t$ denote the state and action at time $t$, respectively. $S_0$ is the initial state.

Deep reinforcement learning extends these concepts by using deep neural networks to approximate the policy $\pi$ or the value functions associated with the states, which help in making decisions. This approach has been pivotal in solving complex decision-making tasks that require high-dimensional state and action spaces.

Tuning of hyperparameters such as the learning rate and the discount factor $\gamma$ plays a critical role in the convergence and performance of RL algorithms. Prominent algorithms like Proximal Policy Optimization (PPO)\cite{schulman2017proximal}, Double Deep Q-Network (DDQN)\cite{mnih2015human}, and Advantage Actor-Critic (A2C)\cite{mnih2016asynchronous} have shown different sensitivities to these parameters. Recent advancements have demonstrated the efficacy of adapting these hyperparameters dynamically in response to ongoing learning progress, which can significantly enhance learning efficiency and effectiveness in complex environments like the Hungry Thirsty Domain. \sys allows users to adjust hyperparameters and incorporate feedback to reduce the time to find an optimal policy without interrupting training. In section \ref{subsection:rewardshaping} we demonstrate the effectiveness of dynamic reward shaping and hyperparameter tuning to for teaching deep reinforcement learning agents.

\subsection{Hungry Thirsty Domain}

We assess our reinforcement learning strategies through reward shaping within the Hungry Thirsty domain \cite{singh2009rewards}. This scenario involves a $4\times 4$ grid, where obstacles block certain paths between adjacent blocks (see Fig \ref{fig}). In randomly assigned corners of the grid, food and water are placed. Following Booth et al. \cite{booth2023perils}, each episode is capped at 200 steps, using a modified environment to increase the challenge.

The agent possesses a simple set of actions: moving in cardinal directions, eating, or drinking. Its primary objectives are to avoid hunger by consuming food and managing thirst, which is compulsory for eating. The agent is classified as not hungry if it has consumed food in the preceding step and can only consume food if not thirsty. To quench its thirst, the agent must be situated on a water tile. Post-drinking, the agent faces a 10\% chance of becoming thirsty again in every consequent step, necessitating a return to water. The agent's state at any time includes its grid location and boolean indicators for hunger \texttt{H} and thirst \texttt{T}.

Performance is evaluated by measuring the agent's fitness, defined as the number of steps spent not hungry, calculated as $F(\tau)=\sum_{t=1}^{200}\mathbb{I}(\neg\texttt{H})$. The agent’s optimal policy $\pi$ directs it towards water when thirsty and towards food otherwise.

The reward function structure is designed to encourage the agent to manage its states effectively:
\begin{equation*}
    \begin{aligned}
        r(\texttt{H} \land \texttt{T}) &= R_{4}, & r(\texttt{H} \land \neg\texttt{T}) &= R_{2}, \\
        r(\neg\texttt{H} \land \texttt{T}) &= R_{3}, & r(\neg\texttt{H} \land \neg\texttt{T}) &= R_{1}.
    \end{aligned}
\end{equation*}
Participants in our experiments adjust these rewards $(R_1, R_2, \\R_3, R_4)$ within the range of [-1, 1]. The challenge is to create a reward scheme that accurately reflects the objective of minimizing hunger, a nontrivial task in reinforcement learning. Booth et al. demonstrate that suboptimal reward structures often result from incorrect assumptions about the spatial relationship between food and water.

Section \ref{subsection:rewardshaping} discusses how experiment participants utilize a heatmap of the most frequent states by the agent (see Fig \ref{fig:heatmap}) to dynamically adjust rewards to optimize fitness. \sys enables continuous, real-time monitoring and adjustment of parameters in training environments such as the Hungry Thirsty domain. This system allows for immediate and ongoing optimizations without the need to pause or restart the training process, thereby enhancing learning efficiency and adaptiveness.

\subsection{Green Computing}
Green computing entails the eco-conscious utilization of computing resources, aiming to maximize energy efficiency while minimizing environmental impact \cite{wang2008meeting}. This is crucial in AI where, for instance, training a natural language processing model can emit approximately 78,000 pounds of CO$_2$—more than double the annual carbon footprint of an individual in the US \cite{strubell2019energy}.

A primary strategy in green computing involves developing processor architectures that enhance performance per watt \cite{zhong2010green}. Such advancements ensure that CPUs and GPUs perform more efficiently, reducing both energy consumption and CO$_2$ emissions. Concurrently, optimizing software algorithms, particularly in deep neural network (DNN) training, can also yield significant efficiencies. By refining these algorithms to converge more rapidly and obviate the need for frequent restarts during training, programs like \sys further reduce energy use and resource waste, thereby shortening the lifecycle of model training.

These approaches collectively strive to sustain technological advancement without the accompanying environmental cost, aligning with the broader objectives of green computing to mitigate the ecological footprint of modern computing technologies.

%% file: 2_related.tex
\begin{figure*}[t]
  \centering
  % \fbox{\rule{0pt}{2in} \rule{0.9\linewidth}{0pt}}
   \includegraphics[width=0.7\textwidth]{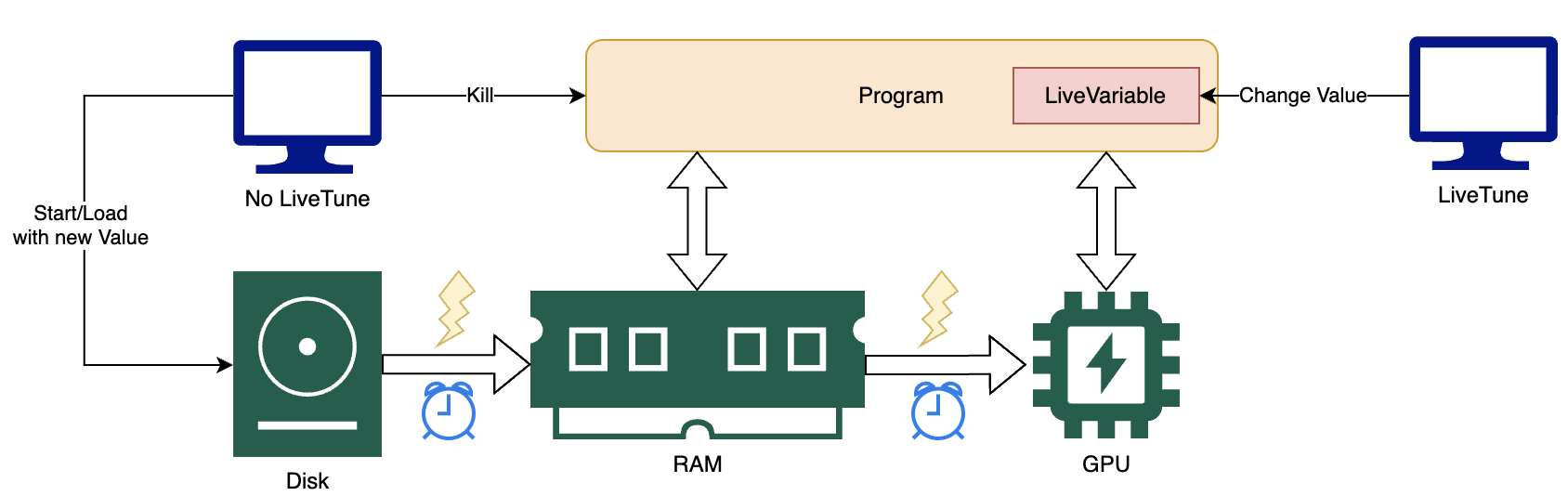}
   \caption{Variable change procedure for \sys and conventional method.}
   \label{fig:main}
\end{figure*}
\vspace{-2mm}
\section{Related Work}
% \nojan{We can add more discussion about RL if there is any - we shouldn't just focus on ML. If there's no RL work we should mention that. Either way, this first paragraph should change to be a little bit more general and introduce why we do or don't mention RL}

Attempts at improving feedback-driven optimization workflows have been most prominent in the area of machine learning. Alongside this, improving reinforcement learning workflows has been recently proposed to allow for easier and faster training procedures. There are several approaches towards increasing efficiency in optimization workflows for these learning paradigms, each targeting different operations within learning workflows to achieve a singular goal: reducing training time. This section reviews key developments in this domain, highlighting their contributions and limitations. We examine the traditional approaches in auto hyperparameter tuning, the use of schedulers for adaptive parameter control, and checkpointing strategies for training robustness. Alongside this, we highlight seminal open problems brought forth in previous works regarding automating RL workflows. These methods, while instrumental in advancing learning workflows, also reveal the need for more dynamic and less resource-intensive solutions.

\noindent\textbf{Auto Hyperparameter Tuning. } Approaches to automate hyperparameter tuning include grid and randomized grid search. These methods can be expedited through prior knowledge or theoretical constraints \cite{smith2018disciplined}. AdaNS, an adaptive genetic-based algorithm introduced by Javaheripi et al., exemplifies this approach \cite{javaheripi2020adans}.

\noindent\textbf{Schedulers. }
Certain hyperparameters, notably the learning rate, often need adjustment during training. For instance, Stochastic Gradient Descent (SGD) benefits from decaying learning rate schedules \cite{ruder2016overview} \cite{zhang2020kdecay} \cite{hsueh2018stochastic}. The Adam optimizer \cite{kingma2014adam}, though adaptively modifying the learning rate based on first and second-moment estimates, is commonly paired with a learning rate scheduler.

\noindent\textbf{Checkpointing. }
Checkpointing is a standard practice in machine learning to safeguard against progress loss due to unexpected performance issues. This approach involves regularly saving model updates. Typically, human experts monitor the training process to determine the optimal time for restarting the model with updated hyperparameters, using a previously saved checkpoint. Recent years have seen a growing interest in researching checkpointing strategies. One of the early contributions in this area was by Vinay et al., who explored the use of MPI (Message Passing Interface) in creating fault-tolerant deep learning (DL) applications, with a focus on checkpoint-restart mechanisms \cite{amatya2017does}. Another significant study by Rojas et al. examined checkpointing in deep neural networks on a large scale. They highlighted a notable overhead issue, identifying significant idle time for GPUs during the checkpointing process \cite{rojas2020study}.

\noindent\textbf{Automating RL. }
AutoRL is an emerging field of research that aims to enable more efficient optimization workflows for RL \cite{parker2022automated}. This work highlights open problems that exist in this space, emphasizing that classical automated hyperparameter tuning in RL is extremely challenging due to the huge hyperparameter search space. The authors list this as an open problem that has not yet been solved.

\sys brings forth a new paradigm to the field of feedback-driven optimization, especially in ML and RL environments. \sys eliminates the need for expensive restarts by providing a framework allowing the change of parameters from outside the running algorithm. Alongside this, \sys can even be used at a higher level to allow a developer to continuously test parameters in learning settings and manually shrink the hyperparameter search space based on feedback, thus lowering the computational complexity of automating optimization for learning tasks.
% , removing the need to load and setup a checkpoint.

% \aiden {we likely want to paint this as a new alternative to checkpoints and restarts. thus, related work should include the papers done on the initial ML process (including checkpoints) here's an example: https://arxiv.org/pdf/2012.00825.pdf}

%% file: 3_method.tex
\section{Methodology}
\label{sec:method}

\subsection{The \sys Framework}
This section details the \sys framework, focusing on its innovative approach to enabling continuous training through the use of LiveVariables. 

\noindent\textbf{LiveVariables:}
LiveVariables are a specialized class of variables designed for dynamic adjustments during runtime without restarting the training process. Upon creation, each LiveVariable allocates a port on the host machine, replacing static variables, such as learning rates, with a dynamic counterpart. These variables are managed by initiating a dedicated listener thread per instance, which monitors and updates the variable's value safely using TCP and semaphore mechanisms. The internal state of a LiveVariable is composed of the current value, a unique tag, and the assigned port, which can be queried through its instance.

\noindent\textbf{Value Modification:}
Modification of a LiveVariable's value during runtime is performed by sending a new value to its listener port via TCP. This is managed safely with semaphores to prevent conflicts. The \texttt{is\_changed()} method is crucial for determining if updates to the variable are necessary, facilitating efficient continuous training.

\noindent\textbf{Dictionary Thread:}
The management of multiple LiveVariable ports is streamlined through a dictionary thread. Employing the singleton pattern \cite{gamma1993design}, this thread maintains a unique instance that logs each LiveVariable's tag and port, ensuring system-wide consistency. Communication with this central dictionary is essential for coordinating updates and maintaining system integrity.

\begin{figure}[t]
  \centering
  % \fbox{\rule{0pt}{2in} \rule{0.9\linewidth}{0pt}}
   \includegraphics[width=0.8\linewidth]{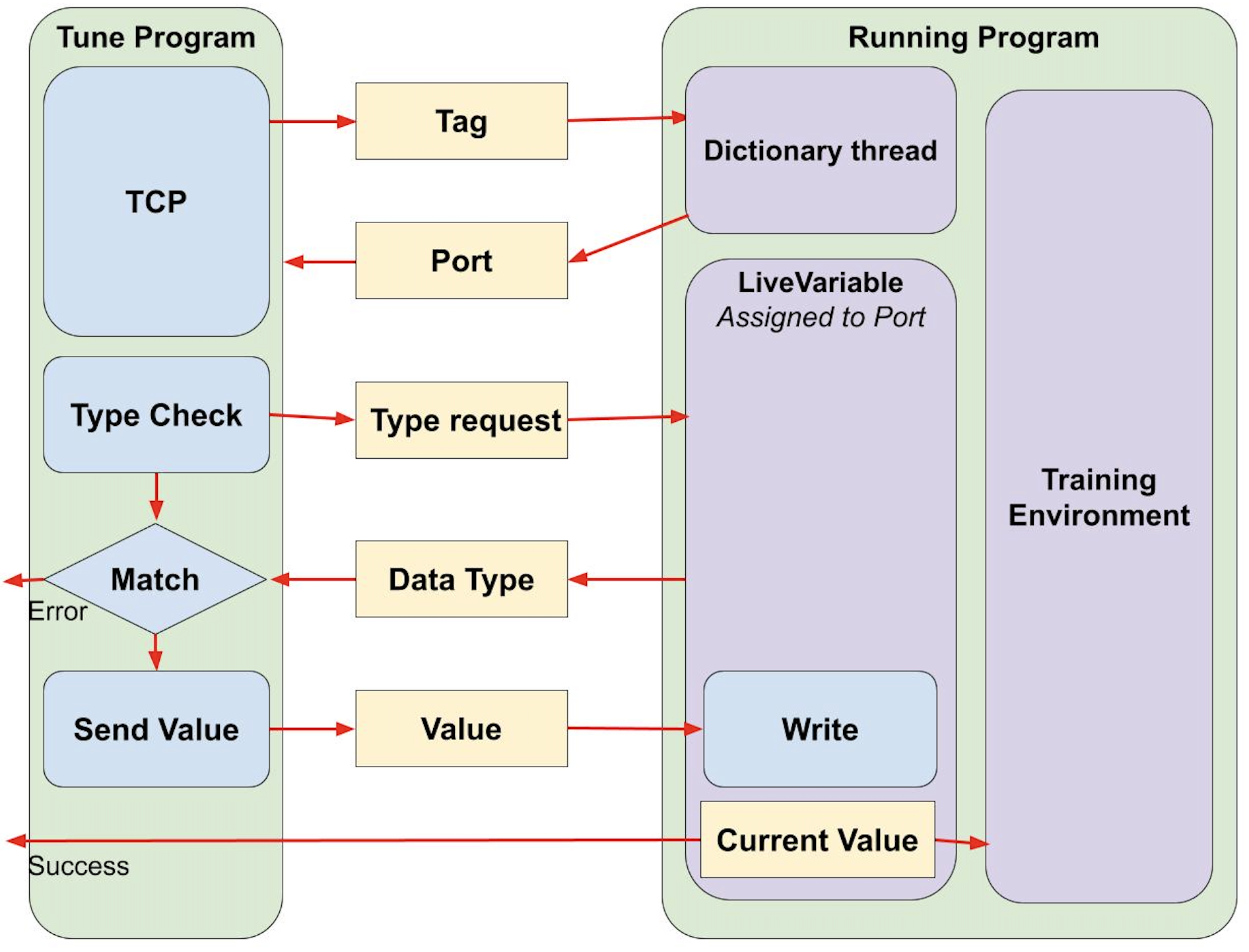}

   \caption{\sys variable change procedure.}
   \label{fig:method}
\end{figure}

\noindent\textbf{Tuning Interface:}
The user interface of the \sys system, referred to as the ``tune'' program, simplifies the process of variable adjustment. It requires the dictionary port, a tag, and a new value for the operation, establishing a secure communication link with the dictionary thread to verify and update values as depicted in Figure \ref{fig:method}. This process ensures that adjustments are applied correctly without type mismatches or disruptions to the main program.

By integrating these components, the \sys framework supports real-time hyperparameter tuning, significantly enhancing the flexibility and efficiency of machine learning training processes.

\subsection{The \sys Workflow}

\sys is a versatile framework designed to accommodate a variety of workflows across different domains. It does not prescribe a specific workflow but instead provides tools that enhance interaction with program variables dynamically through Live Variables. By design, Live Variables are engineered to return their most recent values upon being called, making them ideally suited for use within repetitive loop structures.

In typical use cases, each iteration within a loop provides feedback to the user in the form of log files, output data, or visual plots. An example of this feedback mechanism in action is detailed in Section \ref{subsection:rewardshaping}. Based on this feedback, users can make informed decisions about necessary adjustments to the program's variables. Through the provided API, these changes can be implemented instantly, affecting the subsequent iteration. For instance, in a supervised learning scenario, the learning rate could be set as a Live Variable. As the training progresses, the program outputs the current training and test losses at each iteration. Analyzing these loss curves enables the expert to adjust the learning rate dynamically, optimizing the training process in real time.

This capability is particularly valuable during initial experimentation phases when the optimal settings of hyperparameters are unknown, and rapid iteration is necessary to approximate the best parameters. Once a suitable range of hyperparameter values is identified, it is advisable to transition to more traditional optimization methods, such as grid search or checkpointing, to fine-tune the values and achieve the best performance.

\subsection{Continuous Training}

% \nojan{need to make this more general}

\textbf{Compatibility with Training Loops.} \sys can be integrated with established ML and RL frameworks such as PyTorch and TensorFlow seamlessly. These frameworks typically manage data processing through a loop that iterates over datasets in epochs. \sys facilitates this by enabling checks and resets of the optimizer and other hyperparameters, like batch size, at the start of each epoch. An example of this integration strategy is illustrated in Algorithm \ref{alg:train}, showing how \sys enhances typical training procedures without disrupting existing workflows.

\noindent\textbf{Efficient Hyperparameter Adjustment.} To minimize unnecessary computational overhead during training, \sys employs the \texttt{is\_changed()} method for each LiveVariable. This method efficiently assesses whether adjustments to hyperparameters are required, optimizing resource utilization and system performance. Our experimental results indicate that, while the impact on performance varies, the optimizer resets generally introduce negligible overhead.

In conventional machine learning workflows, any adjustment to a hyperparameter necessitates halting the current program instance and loading a new instance with the updated settings from disk—a process that significantly wastes computational resources \cite{rojas2020study}. In contrast, \sys implements hyperparameter updates directly within the existing program instance, eliminating the need for reloads and thereby reducing the overhead associated with such changes. This method not only saves on computational resources but also reduces the time spent in non-productive setups.

Figure \ref{fig:main} contrasts the workflow of traditional hyperparameter adjustment with that of \sys, highlighting the efficiency improvements by eliminating the need for program restarts and reducing the operational overhead associated with typical hyperparameter adjustments.

\begin{algorithm}[h]
\caption{Continuous Training Loop with Dynamic Learning Rate Adjustment}
\label{alg:generator}
\SetKwProg{generate}{Function \emph{Train}}{}{end}
\While{True}{
    train(model, dataloader, optimizer) \\
    \If{LR.is\_changed()}{
        optimizer $\gets$ Optimizer(LR())
    }
}
\label{alg:train}
\end{algorithm}

%% file: 4_eval.tex
\section{Experimental Evaluation}

In this section, we evaluate our method on both usability and energy consumption. Initially, we explore the energy and time required for restarting a program. We show the effect of model and dataset size in the resources required for a typical machine learning algorithms setup procedure. Subsequently, we conduct two in-depth user studies on \sys's efficiency: 1) two groups compete to train a neural network given limited resources and 2) individual users aim to train an optimal policy in a reinforcement learning task. 
% \nojan{edit}

% Subsequently we give comprehensive insight in an ablation study of \sys's efficiency in a human study, where two groups compete to train a model given limited resources.

\subsection{Energy and Time Consumption}
% \noindent\textbf{Setup. }We utilize a basic and popular image classification algorithm using the pytorch \cite{torchvision2016} framework. To show the effect of model size, we use different variants of the VGG architecture \cite{simonyan2014very}. We used ImageNet-1k \cite{ILSVRC15}, TinyImageNet \cite{le2015tiny}, CIFAR100 \cite{krizhevsky2009learning}. These models are the most widely used image classification datasets. Choice of dataset impacts the loading time. The ImageNet dataset is made of ... 224$\times$224 images. TinyImageNet is composed of ... 64$\times$64 colored images and CIFAR100 comprises of ... 32$\times$32 images. 
% Experiments are done on a single Nvidia A6000 GPU with AMD Ryzen Threadripper 3990X 64-Core Processor and memory speed of 3200 megatransfers per second. All experiments are an average of 6 consecutive runs.

\noindent\textbf{Setup.} Our experiments use a well-known image classification algorithm in the PyTorch framework \cite{torchvision2016}. We examine the impact of model size by employing various versions of the VGG architecture \cite{simonyan2014very}. Three datasets are used: ImageNet-1k \cite{ILSVRC15}, TinyImageNet \cite{le2015tiny}, and CIFAR100 \cite{krizhevsky2009learning}. These are popular choices for image classification tasks.

\begin{itemize}
    \item ImageNet contains large 224x224 pixel images.
    \item TinyImageNet has medium-sized 64x64 pixel images.
    \item CIFAR100 includes small 32x32 pixel images.
    \item Each dataset's size affects how long it takes to load it into the program.
\end{itemize}

We conducted our tests on a powerful computer setup. This includes a single Nvidia A6000 GPU and an AMD Ryzen Threadripper 3990X 64-Core Processor. The memory operates at a speed of 3200 megatransfers per second. We track the execution time of each part of the algorithm, reporting the warmup time when the model starts iterating through batches with a stable flow.To ensure reliability, we repeated each experiment six times and calculated the average results.

% \noindent\textbf{Discussion. } Results for the average loading time and average GPU power consumption can be found in figures \ref{fig:time} and \ref{fig:energy} respectively. As illustrated, there is a clear trend on dataset image size, also implying bigger model size.
% Our algorithm can be broken to four different phases. First, python and library code loading. Each start of a python program induces an overhead for loading the python interpreter from the memory. Secondly, defining the model, dataloader, and the optimizer. Finally loading the model parameters and data batch from disk, then moving the code, data, and parameters to the hardware accelerator. The final phase is the most time and energy consuming phase. In our experiments the GPU setup and data movement took up to 50 seconds and 5.4 Killojoules of wasted energy. 

\begin{figure}[t]
  \centering
  % \fbox{\rule{0pt}{2in} \rule{0.9\linewidth}{0pt}}
   \includegraphics[width=0.7\linewidth]{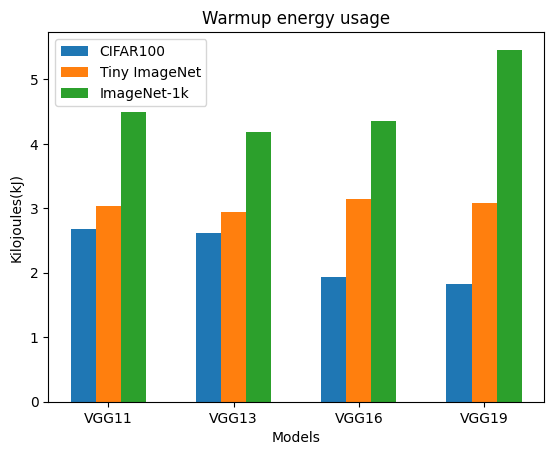}

   \caption{GPU energy consumption of each warmup of our algorithm on different model sizes.}
   \label{fig:energy}
\end{figure}

\begin{figure}[t]
  \centering
  % \fbox{\rule{0pt}{2in} \rule{0.9\linewidth}{0pt}}
   \includegraphics[width=0.7\linewidth]{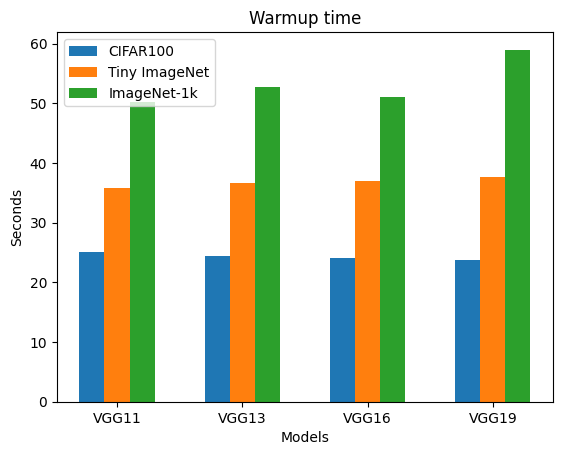}

   \caption{Warmup time of our base algorithm on different model sizes.}
   \label{fig:time}
\end{figure}

\noindent\textbf{Discussion.} Figures \ref{fig:energy} and \ref{fig:time} show the average GPU power consumption and loading time, respectively. These figures reveal a consistent pattern: larger datasets and models tend to take longer to load and consume more power.
We divide the basic classification algorithm into four main stages:
\begin{enumerate}
    \item Python and Library Loading: Starting a Python program requires loading the Python interpreter into memory, which takes time.

    \item Setting Up the Model: This involves defining the model, the data loader, and the optimizer.

    \item Loading Model Parameters and Data: This step reads the model parameters and data batch from the disk.

    \item Moving to the Hardware Accelerator: The most demanding stage in terms of time and energy. Here, we transfer the code, data, and parameters to the GPU.
\end{enumerate}

In our tests, setting up the GPU and transferring data consumed up to 50 seconds and used about 5.4 kJ of energy.

\subsection{Reward Shaping Using \sys}
\label{subsection:rewardshaping}
\textit{Sutton and Barto} \cite{sutton2018reinforcement} emphasize that the reward signal should directly reflect the task goals rather than the methods to achieve them. This principle underlines that a reward function must encode the actual performance metrics of the task. Sparse rewards, which are challenging to learn from, are thus infrequently utilized as highlighted by \textit{Knox et al.} \cite{knox2023reward}. The complexity of selecting an appropriate reward function is well-documented by \textit{Booth et al.} \cite{booth2023perils} through their experiments in the Hungry Thirsty Domain \cite{singh2009rewards}. They gathered 30 experts from top-tier US universities who either actively use reinforcement learning (RL) in their research or have substantial academic exposure to it. For participant details, refer to Appendix A in \cite{booth2023perils}. In their experiments, these experts spent an hour developing optimal reward functions and choosing between algorithms like DDQN \cite{mnih2015human}, PPO \cite{schulman2017proximal}, and A2C \cite{mnih2016asynchronous}, along with setting parameters such as gamma, learning rate, and exploration strategies.

\begin{figure}[t]
  \centering
  % \fbox{\rule{0pt}{2in} \rule{0.9\linewidth}{0pt}}
   \includegraphics[width=0.7\linewidth]{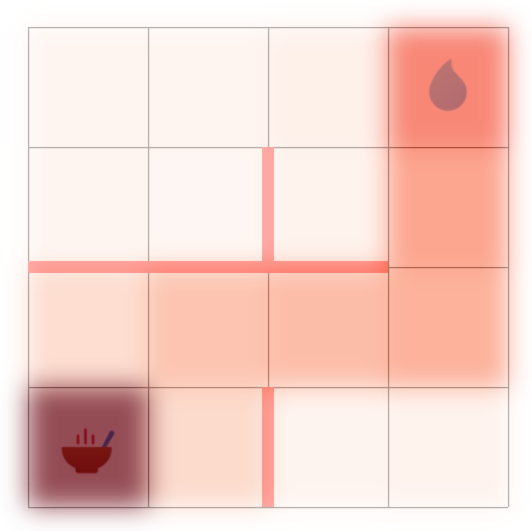}

   \caption{Live heatmap of agent's position on the map. Experts make judgments on how to tune the parameters based on visual feedback.}
   \label{fig:heatmap}
\end{figure}

\begin{figure}[t]
  \centering
  % \fbox{\rule{0pt}{2in} \rule{0.9\linewidth}{0pt}}
   \includegraphics[width=.9\linewidth]{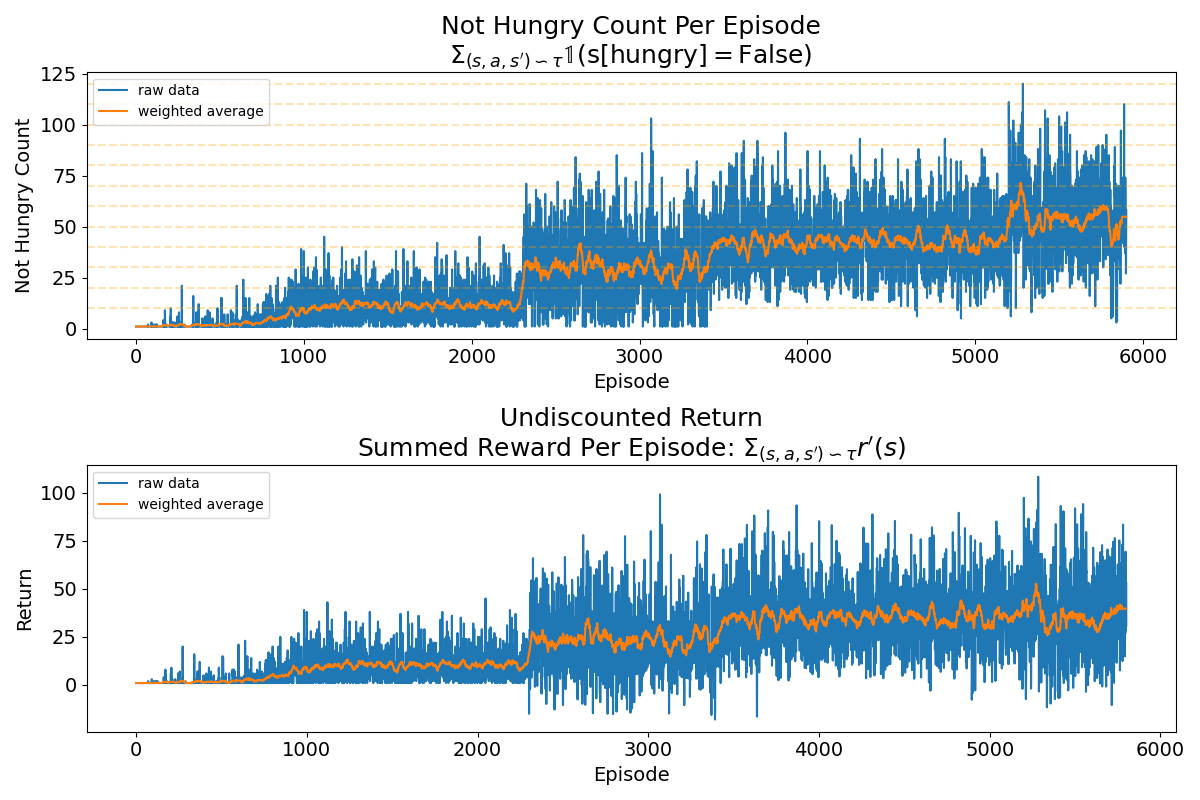}

   \caption{Live plot of the agent's fitness and discounted reward based on the reward function. The jumps observed in the discounted rewards correspond to a \sys-enabled change to the reward function structure by the user.}
   \label{fig:fitness_plot}
\end{figure}

\textbf{Setup.} Our experimental setup mirrored that of \textit{Booth et al.} \cite{booth2023perils}, involving 7 participants with diverse backgrounds, three of whom met the original study’s expertise criteria. The remaining participants, less familiar with RL, received a preliminary tutorial. Utilizing the same algorithms and parameters as the original study, participants aimed to train the most effective RL agent, incorporating \sys for dynamic adjustment of reward functions and algorithm parameters. 
When running the experiment \sys generates a web user interface for users to interact with the program. As the baseline users have access to a live plot of the fitness of their agent and a heat map of how the agent is moving inside the gridworld, shown in figures \ref{fig:heatmap} and \ref{fig:fitness_plot}. Using feedback from the algorithm, the users can change the reward function structure and hyperparameters of the algorithm. All users are evaluated on the time it takes them to teach the agent and the final score they achieved.

\textbf{Discussion.} We utilized the anonymized dataset from Booth et al. for evaluating \sys, comparing it against the traditional method used by \cite{booth2023perils}. Although our participants did not share the same reinforcement learning background as those in the baseline study, the fitness score for agents they trained averaged $86.51$. This contrasts with a baseline average of $5.08$ using the same DDQN algorithm. Our participants reached their final submissions in an average of $35$ minutes, compared to the one hour allowed in \cite{booth2023perils}. In conclusion, the outcomes of this study illustrate the efficacy of \sys in enhancing the performance of reinforcement learning tasks even when participants lack specialized training in this domain. This improvement in agent fitness scores under constrained time conditions, compared to traditional methods, underscores the potential of \sys to facilitate more efficient and effective learning processes.

\subsection{In-Person Competition}

\textbf{Setup.} In our investigation of \sys's applicability for real-time hyperparameter tuning, we organized an in-person competition designed specifically for ML model developers. Eighteen graduate students of varied ML backgrounds participated, grouped into control and experiment groups based on random assignment. Participants faced the challenge of training a deep neural network (DNN) on a subset of the Fashion-MNIST dataset, under constraints that obscured the real hyperparameters by mapping them to four adjustable, but non-linearly correlated, variables. The hyperparameters are learning rate, weight decay, momentum, and dropout rate. The control group managed hyperparameter tuning via a traditional command line interface, necessitating program restarts for each change. Conversely, the experiment group used the \sys framework, facilitating dynamic adjustments through a web interface without needing restarts.

Each participant's machine environment was standardized using Docker, limiting each to 2 processors and 8GB of RAM, thereby negating hardware performance variances. The competition setup also included a Tensorboard interface for immediate feedback on training modifications, emphasizing how changes in hyperparameters impacted model loss and accuracy. The participants were given 1.5 hours to complete their training.

\textbf{Results and Discussion.} The outcome of the competition provided insightful data on \sys’s efficiency. The experiment group, using \sys, achieved an average test accuracy of 85\% by running 135 epochs over the 1.5-hour session. In contrast, the control group reached an 83\% accuracy with 103 epochs. Notably, \sys users conducted on average 33 additional epochs, highlighting the time saved that would otherwise be spent on program restarts and hyperparameter reconfiguration. This efficiency is critical as it suggests a significant reduction in the non-productive overhead associated with traditional hyperparameter tuning.

Furthermore, the experiment underscores \sys's potential to facilitate more environmentally sustainable ML practices by lessening computational waste through efficient resource use. Our findings advocate for broader adoption of \sys in diverse ML training scenarios, potentially revolutionizing hyperparameter tuning practices by significantly curtailing the ecological footprint of ML projects.

% \begin{table}
%     \centering
%     \begin{tabular}{c|cc}
%         \textbf{Rank} & \textbf{Non-LT} & \textbf{LT}\\
%         \midrule
%          &  & \\
%          &  & \\
%          &  & \\
%          &  & \\
%          &  & \\
%     \end{tabular}
%     \caption{Caption}
%     \label{tab:my_label}
% \end{table}

% \nojan{specify what gpu we tested livetune on}

% \subsection{Results}

% \aiden {- experimented LT vs NO LT. lt was way better. - experimented GPU usage changes, this is huge for server clusters. - compared time taken to reach "efficiency", LT takes less time than checkpoints.}

% \section{Discussion}

%% file: 5_conclusion.tex
% \aiden {LiveVariables combined with idle training will further revolutionize ML space, significantly reducing hardware usage and ultimately making ML research more sustainable. this will also save the costs of having to replace parts as often, save energy, and reduce environmental impact.}

\section{Conclusion}

In conclusion, \sys introduces a significant advancement in feedback-driven optimization, particularly in the context of machine learning training and online reward design. This framework allows for real-time dynamic adjustment of parameters, offering a novel approach that enhances machine learning workflows by notably improving efficiency and reducing both time and computational resource consumption. Through the implementation of "LiveVariables" and "LiveTriggers," \sys provides flexible, on-the-fly tuning capabilities without the need for restarting training processes. Our evaluations demonstrate considerable savings, reducing time and energy usage by up to 60 seconds and 5.4 Kilojoules per hyperparameter adjustment, respectively. Furthermore, the application of \sys in reward shaping within reinforcement learning workflows has demonstrated a transformative impact, achieving up to a 5× improvement in efficiency over traditional methods. This enhancement is particularly crucial in domains such as robotics and human-computer interaction, where adaptive learning and quick responsiveness to environmental feedback are essential. By enabling more dynamic and responsive tuning of reinforcement models, \sys not only optimizes learning outcomes but also significantly accelerates the development of more intuitive and interactive robotic systems and HCI interfaces. These advancements promise to drive forward the capabilities of autonomous systems and improve the synergy between humans and technology, underscoring the broad and versatile applicability of \sys in cutting-edge technology sectors. While this work focuses primarily on prominent learning paradigms, \sys can serve as a very valuable tool for any continuous feedback-driven optimization workflow. In future work, we aim to explore the integration of \sys into broader types of computational workflows and further optimization of its core components to support an even wider range of applications, such as CAD optimization, and machine learning paradigms.

%% file: main.bbl
%%% -*-BibTeX-*-
%%% Do NOT edit. File created by BibTeX with style
%%% ACM-Reference-Format-Journals [18-Jan-2012].

\begin{thebibliography}{32}

%%% ====================================================================
%%% NOTE TO THE USER: you can override these defaults by providing
%%% customized versions of any of these macros before the \bibliography
%%% command.  Each of them MUST provide its own final punctuation,
%%% except for \shownote{}, \showDOI{}, and \showURL{}.  The latter two
%%% do not use final punctuation, in order to avoid confusing it with
%%% the Web address.
%%%
%%% To suppress output of a particular field, define its macro to expand
%%% to an empty string, or better, \unskip, like this:
%%%
%%% \newcommand{\showDOI}[1]{\unskip}   % LaTeX syntax
%%%
%%% \def \showDOI #1{\unskip}           % plain TeX syntax
%%%
%%% ====================================================================

\ifx \showCODEN    \undefined \def \showCODEN     #1{\unskip}     \fi
\ifx \showDOI      \undefined \def \showDOI       #1{#1}\fi
\ifx \showISBNx    \undefined \def \showISBNx     #1{\unskip}     \fi
\ifx \showISBNxiii \undefined \def \showISBNxiii  #1{\unskip}     \fi
\ifx \showISSN     \undefined \def \showISSN      #1{\unskip}     \fi
\ifx \showLCCN     \undefined \def \showLCCN      #1{\unskip}     \fi
\ifx \shownote     \undefined \def \shownote      #1{#1}          \fi
\ifx \showarticletitle \undefined \def \showarticletitle #1{#1}   \fi
\ifx \showURL      \undefined \def \showURL       {\relax}        \fi
% The following commands are used for tagged output and should be
% invisible to TeX
\providecommand\bibfield[2]{#2}
\providecommand\bibinfo[2]{#2}
\providecommand\natexlab[1]{#1}
\providecommand\showeprint[2][]{arXiv:#2}

\bibitem[Amatya et~al\mbox{.}(2017)]%
        {amatya2017does}
\bibfield{author}{\bibinfo{person}{Vinay Amatya}, \bibinfo{person}{Abhinav Vishnu}, \bibinfo{person}{Charles Siegel}, {and} \bibinfo{person}{Jeff Daily}.} \bibinfo{year}{2017}\natexlab{}.
\newblock \showarticletitle{What does fault tolerant deep learning need from mpi?}. In \bibinfo{booktitle}{\emph{Proceedings of the 24th European MPI Users' Group Meeting}}. \bibinfo{pages}{1--11}.
\newblock


\bibitem[Bardenet et~al\mbox{.}(2013)]%
        {bardenet2013collaborative}
\bibfield{author}{\bibinfo{person}{R{\'e}mi Bardenet}, \bibinfo{person}{M{\'a}ty{\'a}s Brendel}, \bibinfo{person}{Bal{\'a}zs K{\'e}gl}, {and} \bibinfo{person}{Michele Sebag}.} \bibinfo{year}{2013}\natexlab{}.
\newblock \showarticletitle{Collaborative hyperparameter tuning}. In \bibinfo{booktitle}{\emph{International conference on machine learning}}. PMLR, \bibinfo{pages}{199--207}.
\newblock


\bibitem[Booth et~al\mbox{.}(2023)]%
        {booth2023perils}
\bibfield{author}{\bibinfo{person}{Serena Booth}, \bibinfo{person}{W~Bradley Knox}, \bibinfo{person}{Julie Shah}, \bibinfo{person}{Scott Niekum}, \bibinfo{person}{Peter Stone}, {and} \bibinfo{person}{Alessandro Allievi}.} \bibinfo{year}{2023}\natexlab{}.
\newblock \showarticletitle{The perils of trial-and-error reward design: misdesign through overfitting and invalid task specifications}. In \bibinfo{booktitle}{\emph{Proceedings of the AAAI Conference on Artificial Intelligence}}, Vol.~\bibinfo{volume}{37}. \bibinfo{pages}{5920--5929}.
\newblock


\bibitem[Gamma et~al\mbox{.}(1993)]%
        {gamma1993design}
\bibfield{author}{\bibinfo{person}{Erich Gamma}, \bibinfo{person}{Richard Helm}, \bibinfo{person}{Ralph Johnson}, {and} \bibinfo{person}{John Vlissides}.} \bibinfo{year}{1993}\natexlab{}.
\newblock \showarticletitle{Design patterns: Abstraction and reuse of object-oriented design}. In \bibinfo{booktitle}{\emph{ECOOP’93—Object-Oriented Programming: 7th European Conference Kaiserslautern, Germany, July 26--30, 1993 Proceedings 7}}. Springer, \bibinfo{pages}{406--431}.
\newblock


\bibitem[Hsueh et~al\mbox{.}(2018)]%
        {hsueh2018stochastic}
\bibfield{author}{\bibinfo{person}{Bo~Yang Hsueh}, \bibinfo{person}{Wei Li}, {and} \bibinfo{person}{I-Chen Wu}.} \bibinfo{year}{2018}\natexlab{}.
\newblock \showarticletitle{Stochastic gradient descent with hyperbolic-tangent decay}.
\newblock \bibinfo{journal}{\emph{CoRR abs/1806.01593}} (\bibinfo{year}{2018}).
\newblock


\bibitem[Javaheripi et~al\mbox{.}(2020)]%
        {javaheripi2020adans}
\bibfield{author}{\bibinfo{person}{Mojan Javaheripi}, \bibinfo{person}{Mohammad Samragh}, \bibinfo{person}{Tara Javidi}, {and} \bibinfo{person}{Farinaz Koushanfar}.} \bibinfo{year}{2020}\natexlab{}.
\newblock \showarticletitle{AdaNS: Adaptive non-uniform sampling for automated design of compact DNNs}.
\newblock \bibinfo{journal}{\emph{IEEE Journal of Selected Topics in Signal Processing}} \bibinfo{volume}{14}, \bibinfo{number}{4} (\bibinfo{year}{2020}), \bibinfo{pages}{750--764}.
\newblock


\bibitem[Joy et~al\mbox{.}(2016)]%
        {joy2016hyperparameter}
\bibfield{author}{\bibinfo{person}{Tinu~Theckel Joy}, \bibinfo{person}{Santu Rana}, \bibinfo{person}{Sunil Gupta}, {and} \bibinfo{person}{Svetha Venkatesh}.} \bibinfo{year}{2016}\natexlab{}.
\newblock \showarticletitle{Hyperparameter tuning for big data using Bayesian optimisation}. In \bibinfo{booktitle}{\emph{2016 23rd International Conference on Pattern Recognition (ICPR)}}. IEEE, \bibinfo{pages}{2574--2579}.
\newblock


\bibitem[Kingma and Ba(2014)]%
        {kingma2014adam}
\bibfield{author}{\bibinfo{person}{Diederik~P Kingma} {and} \bibinfo{person}{Jimmy Ba}.} \bibinfo{year}{2014}\natexlab{}.
\newblock \showarticletitle{Adam: A method for stochastic optimization}.
\newblock \bibinfo{journal}{\emph{arXiv preprint arXiv:1412.6980}} (\bibinfo{year}{2014}).
\newblock


\bibitem[Knox et~al\mbox{.}(2023)]%
        {knox2023reward}
\bibfield{author}{\bibinfo{person}{W~Bradley Knox}, \bibinfo{person}{Alessandro Allievi}, \bibinfo{person}{Holger Banzhaf}, \bibinfo{person}{Felix Schmitt}, {and} \bibinfo{person}{Peter Stone}.} \bibinfo{year}{2023}\natexlab{}.
\newblock \showarticletitle{Reward (mis) design for autonomous driving}.
\newblock \bibinfo{journal}{\emph{Artificial Intelligence}}  \bibinfo{volume}{316} (\bibinfo{year}{2023}), \bibinfo{pages}{103829}.
\newblock


\bibitem[Krizhevsky et~al\mbox{.}(2009)]%
        {krizhevsky2009learning}
\bibfield{author}{\bibinfo{person}{Alex Krizhevsky}, \bibinfo{person}{Geoffrey Hinton}, {et~al\mbox{.}}} \bibinfo{year}{2009}\natexlab{}.
\newblock \showarticletitle{Learning multiple layers of features from tiny images}.
\newblock  (\bibinfo{year}{2009}).
\newblock


\bibitem[Le and Yang(2015)]%
        {le2015tiny}
\bibfield{author}{\bibinfo{person}{Ya Le} {and} \bibinfo{person}{Xuan Yang}.} \bibinfo{year}{2015}\natexlab{}.
\newblock \showarticletitle{Tiny imagenet visual recognition challenge}.
\newblock \bibinfo{journal}{\emph{CS 231N}} \bibinfo{volume}{7}, \bibinfo{number}{7} (\bibinfo{year}{2015}), \bibinfo{pages}{3}.
\newblock


\bibitem[maintainers and contributors(2016)]%
        {torchvision2016}
\bibfield{author}{\bibinfo{person}{TorchVision maintainers} {and} \bibinfo{person}{contributors}.} \bibinfo{year}{2016}\natexlab{}.
\newblock \bibinfo{booktitle}{\emph{TorchVision: PyTorch's Computer Vision library}}.
\newblock


\bibitem[Mantovani et~al\mbox{.}(2015)]%
        {mantovani2015effectiveness}
\bibfield{author}{\bibinfo{person}{Rafael~G Mantovani}, \bibinfo{person}{Andr{\'e}~LD Rossi}, \bibinfo{person}{Joaquin Vanschoren}, \bibinfo{person}{Bernd Bischl}, {and} \bibinfo{person}{Andr{\'e}~CPLF De~Carvalho}.} \bibinfo{year}{2015}\natexlab{}.
\newblock \showarticletitle{Effectiveness of random search in SVM hyper-parameter tuning}. In \bibinfo{booktitle}{\emph{2015 international joint conference on neural networks (IJCNN)}}. Ieee, \bibinfo{pages}{1--8}.
\newblock


\bibitem[Mnih et~al\mbox{.}(2016)]%
        {mnih2016asynchronous}
\bibfield{author}{\bibinfo{person}{Volodymyr Mnih}, \bibinfo{person}{Adria~Puigdomenech Badia}, \bibinfo{person}{Mehdi Mirza}, \bibinfo{person}{Alex Graves}, \bibinfo{person}{Timothy Lillicrap}, \bibinfo{person}{Tim Harley}, \bibinfo{person}{David Silver}, {and} \bibinfo{person}{Koray Kavukcuoglu}.} \bibinfo{year}{2016}\natexlab{}.
\newblock \showarticletitle{Asynchronous methods for deep reinforcement learning}. In \bibinfo{booktitle}{\emph{International conference on machine learning}}. PMLR, \bibinfo{pages}{1928--1937}.
\newblock


\bibitem[Mnih et~al\mbox{.}(2015)]%
        {mnih2015human}
\bibfield{author}{\bibinfo{person}{Volodymyr Mnih}, \bibinfo{person}{Koray Kavukcuoglu}, \bibinfo{person}{David Silver}, \bibinfo{person}{Andrei~A Rusu}, \bibinfo{person}{Joel Veness}, \bibinfo{person}{Marc~G Bellemare}, \bibinfo{person}{Alex Graves}, \bibinfo{person}{Martin Riedmiller}, \bibinfo{person}{Andreas~K Fidjeland}, \bibinfo{person}{Georg Ostrovski}, {et~al\mbox{.}}} \bibinfo{year}{2015}\natexlab{}.
\newblock \showarticletitle{Human-level control through deep reinforcement learning}.
\newblock \bibinfo{journal}{\emph{nature}} \bibinfo{volume}{518}, \bibinfo{number}{7540} (\bibinfo{year}{2015}), \bibinfo{pages}{529--533}.
\newblock


\bibitem[Parker-Holder et~al\mbox{.}(2022)]%
        {parker2022automated}
\bibfield{author}{\bibinfo{person}{Jack Parker-Holder}, \bibinfo{person}{Raghu Rajan}, \bibinfo{person}{Xingyou Song}, \bibinfo{person}{Andr{\'e} Biedenkapp}, \bibinfo{person}{Yingjie Miao}, \bibinfo{person}{Theresa Eimer}, \bibinfo{person}{Baohe Zhang}, \bibinfo{person}{Vu Nguyen}, \bibinfo{person}{Roberto Calandra}, \bibinfo{person}{Aleksandra Faust}, {et~al\mbox{.}}} \bibinfo{year}{2022}\natexlab{}.
\newblock \showarticletitle{Automated reinforcement learning (autorl): A survey and open problems}.
\newblock \bibinfo{journal}{\emph{Journal of Artificial Intelligence Research}}  \bibinfo{volume}{74} (\bibinfo{year}{2022}), \bibinfo{pages}{517--568}.
\newblock


\bibitem[Rodriguez(2018)]%
        {rodriguez2018understanding}
\bibfield{author}{\bibinfo{person}{Jesus Rodriguez}.} \bibinfo{year}{2018}\natexlab{}.
\newblock \showarticletitle{Understanding hyperparameters optimization in deep learning models: concepts and tools}.
\newblock \bibinfo{journal}{\emph{Linkedin Pulse}} (\bibinfo{year}{2018}).
\newblock


\bibitem[Rojas et~al\mbox{.}(2020)]%
        {rojas2020study}
\bibfield{author}{\bibinfo{person}{Elvis Rojas}, \bibinfo{person}{Albert~Njoroge Kahira}, \bibinfo{person}{Esteban Meneses}, \bibinfo{person}{Leonardo~Bautista Gomez}, {and} \bibinfo{person}{Rosa~M Badia}.} \bibinfo{year}{2020}\natexlab{}.
\newblock \showarticletitle{A study of checkpointing in large scale training of deep neural networks}.
\newblock \bibinfo{journal}{\emph{arXiv preprint arXiv:2012.00825}} (\bibinfo{year}{2020}).
\newblock


\bibitem[Ruder(2016)]%
        {ruder2016overview}
\bibfield{author}{\bibinfo{person}{Sebastian Ruder}.} \bibinfo{year}{2016}\natexlab{}.
\newblock \showarticletitle{An overview of gradient descent optimization algorithms}.
\newblock \bibinfo{journal}{\emph{arXiv preprint arXiv:1609.04747}} (\bibinfo{year}{2016}).
\newblock


\bibitem[Russakovsky et~al\mbox{.}(2015)]%
        {ILSVRC15}
\bibfield{author}{\bibinfo{person}{Olga Russakovsky}, \bibinfo{person}{Jia Deng}, \bibinfo{person}{Hao Su}, \bibinfo{person}{Jonathan Krause}, \bibinfo{person}{Sanjeev Satheesh}, \bibinfo{person}{Sean Ma}, \bibinfo{person}{Zhiheng Huang}, \bibinfo{person}{Andrej Karpathy}, \bibinfo{person}{Aditya Khosla}, \bibinfo{person}{Michael Bernstein}, \bibinfo{person}{Alexander~C. Berg}, {and} \bibinfo{person}{Li Fei-Fei}.} \bibinfo{year}{2015}\natexlab{}.
\newblock \showarticletitle{{ImageNet Large Scale Visual Recognition Challenge}}.
\newblock \bibinfo{journal}{\emph{International Journal of Computer Vision (IJCV)}} \bibinfo{volume}{115}, \bibinfo{number}{3} (\bibinfo{year}{2015}), \bibinfo{pages}{211--252}.
\newblock
\urldef\tempurl%
\url{https://doi.org/10.1007/s11263-015-0816-y}
\showDOI{\tempurl}


\bibitem[Schulman et~al\mbox{.}(2017)]%
        {schulman2017proximal}
\bibfield{author}{\bibinfo{person}{John Schulman}, \bibinfo{person}{Filip Wolski}, \bibinfo{person}{Prafulla Dhariwal}, \bibinfo{person}{Alec Radford}, {and} \bibinfo{person}{Oleg Klimov}.} \bibinfo{year}{2017}\natexlab{}.
\newblock \showarticletitle{Proximal policy optimization algorithms}.
\newblock \bibinfo{journal}{\emph{arXiv preprint arXiv:1707.06347}} (\bibinfo{year}{2017}).
\newblock


\bibitem[Shekar and Dagnew(2019)]%
        {shekar2019grid}
\bibfield{author}{\bibinfo{person}{BH Shekar} {and} \bibinfo{person}{Guesh Dagnew}.} \bibinfo{year}{2019}\natexlab{}.
\newblock \showarticletitle{Grid search-based hyperparameter tuning and classification of microarray cancer data}. In \bibinfo{booktitle}{\emph{2019 second international conference on advanced computational and communication paradigms (ICACCP)}}. IEEE, \bibinfo{pages}{1--8}.
\newblock


\bibitem[Simonyan and Zisserman(2014)]%
        {simonyan2014very}
\bibfield{author}{\bibinfo{person}{Karen Simonyan} {and} \bibinfo{person}{Andrew Zisserman}.} \bibinfo{year}{2014}\natexlab{}.
\newblock \showarticletitle{Very deep convolutional networks for large-scale image recognition}.
\newblock \bibinfo{journal}{\emph{arXiv preprint arXiv:1409.1556}} (\bibinfo{year}{2014}).
\newblock


\bibitem[Singh et~al\mbox{.}(2009)]%
        {singh2009rewards}
\bibfield{author}{\bibinfo{person}{Satinder Singh}, \bibinfo{person}{Richard~L Lewis}, {and} \bibinfo{person}{Andrew~G Barto}.} \bibinfo{year}{2009}\natexlab{}.
\newblock \showarticletitle{Where do rewards come from}. In \bibinfo{booktitle}{\emph{Proceedings of the annual conference of the cognitive science society}}. Cognitive Science Society, \bibinfo{pages}{2601--2606}.
\newblock


\bibitem[Smith(2018)]%
        {smith2018disciplined}
\bibfield{author}{\bibinfo{person}{Leslie~N Smith}.} \bibinfo{year}{2018}\natexlab{}.
\newblock \showarticletitle{A disciplined approach to neural network hyper-parameters: Part 1--learning rate, batch size, momentum, and weight decay}.
\newblock \bibinfo{journal}{\emph{arXiv preprint arXiv:1803.09820}} (\bibinfo{year}{2018}).
\newblock


\bibitem[Strubell et~al\mbox{.}(2019)]%
        {strubell2019energy}
\bibfield{author}{\bibinfo{person}{Emma Strubell}, \bibinfo{person}{Ananya Ganesh}, {and} \bibinfo{person}{Andrew McCallum}.} \bibinfo{year}{2019}\natexlab{}.
\newblock \showarticletitle{Energy and policy considerations for deep learning in NLP}.
\newblock \bibinfo{journal}{\emph{arXiv preprint arXiv:1906.02243}} (\bibinfo{year}{2019}).
\newblock


\bibitem[Sutton and Barto(2018)]%
        {sutton2018reinforcement}
\bibfield{author}{\bibinfo{person}{Richard~S Sutton} {and} \bibinfo{person}{Andrew~G Barto}.} \bibinfo{year}{2018}\natexlab{}.
\newblock \bibinfo{booktitle}{\emph{Reinforcement learning: An introduction}}.
\newblock \bibinfo{publisher}{MIT press}.
\newblock


\bibitem[Tan and Le(2019)]%
        {tan2019efficientnet}
\bibfield{author}{\bibinfo{person}{Mingxing Tan} {and} \bibinfo{person}{Quoc Le}.} \bibinfo{year}{2019}\natexlab{}.
\newblock \showarticletitle{Efficientnet: Rethinking model scaling for convolutional neural networks}. In \bibinfo{booktitle}{\emph{International conference on machine learning}}. PMLR, \bibinfo{pages}{6105--6114}.
\newblock


\bibitem[Wang(2008)]%
        {wang2008meeting}
\bibfield{author}{\bibinfo{person}{David Wang}.} \bibinfo{year}{2008}\natexlab{}.
\newblock \showarticletitle{Meeting green computing challenges}. In \bibinfo{booktitle}{\emph{2008 10th Electronics Packaging Technology Conference}}. IEEE, \bibinfo{pages}{121--126}.
\newblock


\bibitem[Yu and Zhu(2020)]%
        {yu2020hyper}
\bibfield{author}{\bibinfo{person}{Tong Yu} {and} \bibinfo{person}{Hong Zhu}.} \bibinfo{year}{2020}\natexlab{}.
\newblock \showarticletitle{Hyper-parameter optimization: A review of algorithms and applications}.
\newblock \bibinfo{journal}{\emph{arXiv preprint arXiv:2003.05689}} (\bibinfo{year}{2020}).
\newblock


\bibitem[Zhang and Li(2020)]%
        {zhang2020kdecay}
\bibfield{author}{\bibinfo{person}{Tao Zhang} {and} \bibinfo{person}{Wei Li}.} \bibinfo{year}{2020}\natexlab{}.
\newblock \showarticletitle{kDecay: Just adding k-decay items on Learning-Rate Schedule to improve Neural Networks}.
\newblock \bibinfo{journal}{\emph{arXiv preprint arXiv:2004.05909}} (\bibinfo{year}{2020}).
\newblock


\bibitem[Zhong et~al\mbox{.}(2010)]%
        {zhong2010green}
\bibfield{author}{\bibinfo{person}{Benjamin Zhong}, \bibinfo{person}{Ming Feng}, {and} \bibinfo{person}{Chung-Horng Lung}.} \bibinfo{year}{2010}\natexlab{}.
\newblock \showarticletitle{A green computing based architecture comparison and analysis}. In \bibinfo{booktitle}{\emph{2010 IEEE/ACM Int'l Conference on Green Computing and Communications \& Int'l Conference on Cyber, Physical and Social Computing}}. IEEE, \bibinfo{pages}{386--391}.
\newblock


\end{thebibliography}
